\documentclass[runningheads]{llncs}
\usepackage[T1]{fontenc}
\usepackage{graphicx}
\usepackage{amssymb}
\usepackage{amsmath}
\usepackage{subfigure}
\usepackage{booktabs}

\usepackage[pagebackref=true,breaklinks=true,letterpaper=true,colorlinks,citecolor=blue,linkcolor=blue,bookmarks=false]{hyperref}
\usepackage{multirow}

\begin{document}
\title{FD-Vision Mamba for Endoscopic Exposure Correction}
\author{Zhuoran Zheng\inst{1} \and
Jun Zhang\inst{2} 
}
%
%
\institute{Nanjing University of Science and Technology \and
Chinese Academy of Sciences \\
\email{zhengzr@njust.edu.cn}}
%
%
%
%
\maketitle              
\begin{abstract}
In endoscopic imaging, the recorded images are prone to exposure abnormalities, so maintaining high-quality images is important to assist healthcare professionals in performing decision-making. 
To overcome this issue, We design a frequency-domain based network, called FD-Vision Mamba (FDVM-Net), which achieves high-quality image exposure correction by reconstructing the frequency domain of endoscopic images.
Specifically, inspired by the State Space Sequence Models (SSMs), we develop a C-SSM block that integrates the local feature extraction ability of the convolutional layer with the ability of the SSM to capture long-range dependencies.
A two-path network is built using C-SSM as the basic function cell, and these two paths deal with the phase and amplitude information of the image, respectively.
Finally, a degraded endoscopic image is reconstructed by FDVM-Net to obtain a high-quality clear image.
Extensive experimental results demonstrate that our method achieves state-of-the-art results in terms of speed and accuracy, and it is noteworthy that our method can enhance endoscopic images of arbitrary resolution.
The URL of the code is \url{https://github.com/zzr-idam/FDVM-Net}.

\keywords{Endoscopic imaging  \and Frequency domain \and FD-Vision Mamba.}
\end{abstract}
\section{Introduction}
%
%
Using an endoscope to check the interior of the human body became a necessity for doctors~\cite{gong2000wireless,iddan2000wireless}.
%
However, endoscopic imaging is often prone to overexposure or underexposure due to the point source ~\cite{sang2020inferring} effect, which largely interferes with the decision-making of healthcare professionals.

%
Recently, deep learning-based methods~\cite{Ma2021StructureAI,Ma2020CycleSA,Zhong2023MultiScaleAG,garcia2023multi,wang2022endoscopic,an2022eien} have marked progress in enhancing endoscopic imagery. 
Although these methods achieve state-of-the-art results on an endoscopic image enhancement task, they generate huge computational losses to model the long-range dependence of the image.
Therefore, how to effectively model long-range dependencies of images still is a challenge on resource-constrained devices. 
Recently, State Space Sequence Models (SSMs)~\cite{GuThesis,SSM-NeurIPS21}, especially Structured State Space Sequence Models (S4) ~\cite{S4-ICLR22}, have emerged as efficient building blocks (or layers) for constructing deep networks, and have gained promising performances in the analysis of continuous long sequence data~\cite{SSM4Audio,S4-ICLR22}.
Mamba~\cite{mamba} further improved S4 using a selection mechanism to outperform Transformers on dense modalities such as language and genomics. In addition, state-space models have shown promising results on visual tasks such as images~\cite{S4ND-neurips22} and videos~\cite{SSM4Video-ECCV22} classification. Since image patches and image features can be used as sequences~\cite{ViT2020,Swin}, these appealing features of state-space models motivate us to explore the use of Mamba blocks to design an efficient model with long-range modeling capabilities.


Inspired by this, we present a novel approach (FDVM-Net) based on the Mamba to effectively reconstruct the exposure of endoscopic images.
Furthermore, we consider the global information characteristic of the phase and amplitude of the frequency domain information~\cite{zhou2023fourmer}, and we try to introduce this principle to integrate with Mamba for quality global modeling.
Specifically, we use Mamba and convolutional blocks (C-SSM) to build a basic cell, and then build a dual-path network based on this basic cell.
This dual-path network is fed with the phase and amplitude information of the image respectively, in which we propose a novel frequency-domain cross-attention to boost the performance of the model.
The purpose of frequency-domain cross-attention is to mutually prompt phase and amplitude which information values are preserved and which information can be overlooked.
Extensive experimental results show that our method performs well in endoscopic image exposure correction tasks.
FDVM-Net paves the way for future designs of medical image enhancement, which can also show more capability in other image restoration tasks.

\begin{figure}[t]
	\begin{center}\scriptsize
		\tabcolsep 1pt
		\begin{tabular}{@{}c@{}}
                \includegraphics[width = 0.95\textwidth]{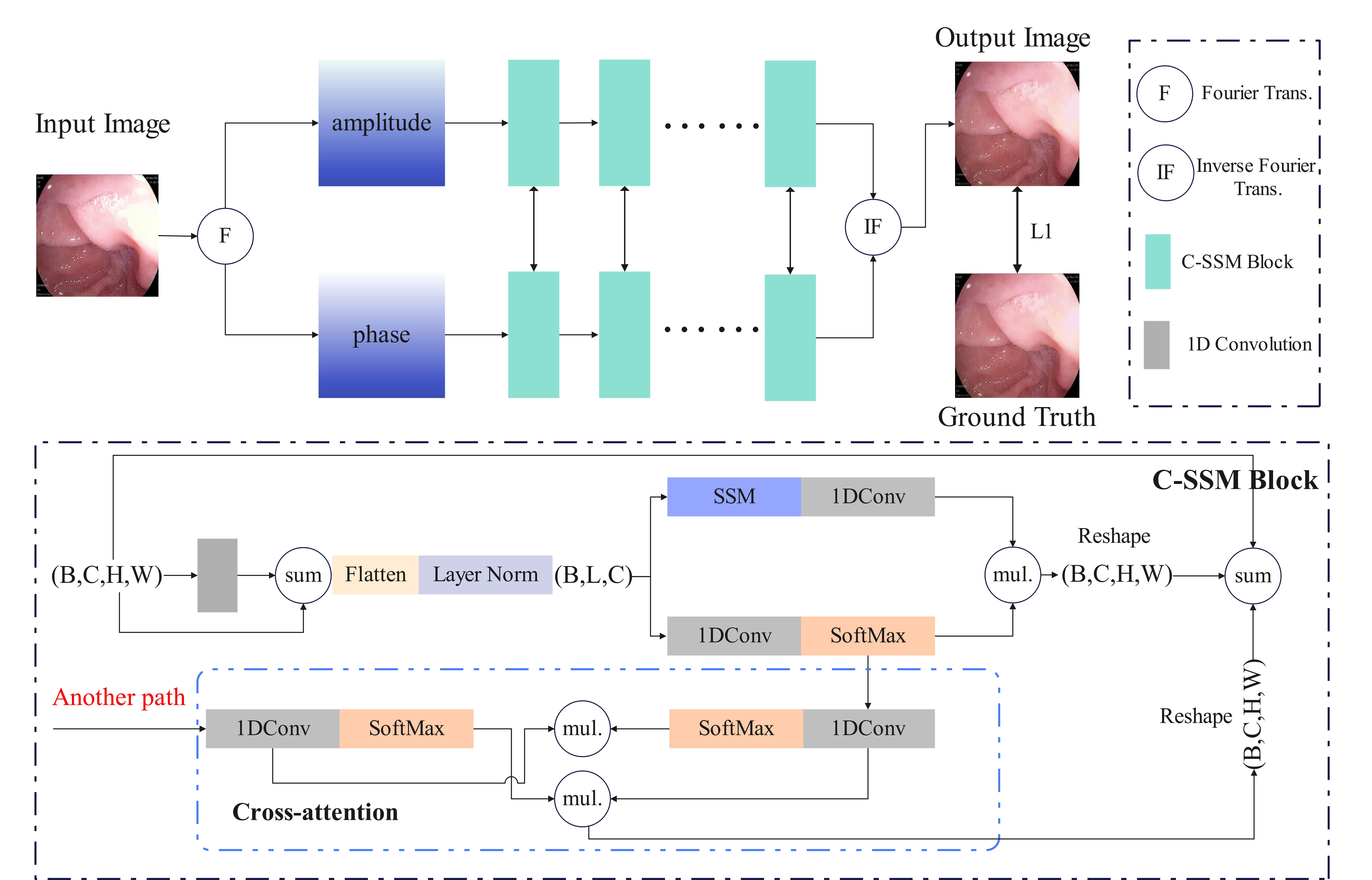}  
		\end{tabular}
	\end{center}
 \vspace{-4mm}
 	\caption{\textbf{Overview of the DFVM-Net architecture.} DFVM-Net is a two-path network with some C-SSM blocks in series with each path. In DFVM-Net block, we use convolution, SSM, cross-attention, and shortcut to form a basic cell. In this network, the upper branch deals with amplitude, the lower branch deals with phase, and finally the feature map is inverse Fourier transformed to yield a clear endoscopic image.}
	\label{f2}
 \vspace{-4mm}
\end{figure}

\section{Method}
Fig.~\ref{f2} shows an overview of the DFVM-Net block and the whole network architecture. Next, we first introduce the Mamba block followed by illustrating the details of DFVM-Net.

\subsection{DFVM-Net}
%
%
%
%
%
As shown in Figure~\ref{f2}, our network is a dual-path architecture.
%
Given an input image $\mathbf{X}$, we obtain the phase $\mathbf{P}$ and amplitude $\mathbf{A}$ after transforming it by Fourier transform. Next, input $\mathbf{P}$ and $\mathbf{A}$ into two paths.

\noindent \textbf{C-SSM block.}
In the C-SSM block, the feature maps ($\mathbf{F}$) start to be fed to the standard convolutional layer ($3 \times 3$ convolutional kernel) with the ReLU.
%
Next, image features with a shape of $(B, C, H, W)$ are then flattened and transposed to $(B, L, C)$ where $L=H\times W$. After passing the Layer Normalization~\cite{layerNorm} (normalization in horizontal features), the features enter two parallel branches. In the first branch, the features are expanded to $(B, L, C)$ by an SSM layer~\cite{mamba} with 1D convolution ($1 \times 3$ convolution kernel). 
In the second branch, the features are also expanded to $(B, L, C)$ by a 1D convolution ($1 \times 3$ convolution kernel), next, the feature map goes through softmax to form an attention map. Then, the features from the two branches are merged with the Hadamard product. Finally, the features are projected back to the original shape $(B, L, C)$ and then reshaped and transposed to $(B, C, H, W)$.
%
%
Note that due to the high computational complexity of SSM, we use bilinear interpolation to fix to a low-resolution feature map ($C \times 64 \times 64$ or $C \times 56 \times 56$) input to SSM, after which the bilinear interpolation back to the original resolution.

\noindent \textbf{Cross-attention.}
In the second branch, the attention map is further extracted by 1D convolution and Softmax to extract feature inputs into another side-by-side C-SSM block of the network.
Then, the features from the two paths are merged with the Hadamard product. Finally, the features are projected back to the original shape $(B, L, C)$ and then reshaped and transposed to $(B, C, H, W)$.
In this paper, 8 C-SSM blocks with cross-attention and shortcuts are used for each path.
For the loss function, only the L1 norm is employed.

\begin{table*}[t] \scriptsize
\centering
\begin{tabular*}{\textwidth}{@{\extracolsep{\fill}}lccccccc} 
\toprule
& LIME \cite{guo2016lime} & HDRNET \cite{gharbi2017deep} & SwinIR \cite{liang2021swinir} & LECCM \cite{nsampi2021learning} & NAFNet \cite{chen2022simple} & EndoIMLE \cite{wang2022endoscopic} & Ours \\
\midrule
PSNR (dB) & 19.18 & 29.65 & 23.69 & 24.63 & 28.85 & 17.43 & \textbf{33.99} \\
SSIM & 0.9447 & \textbf{0.9997} & 0.8881 & 0.9885 & 0.9775 & 0.8727 & 0.9805 \\
Time(ms) & 1670.15 & \textbf{2.46} & 5962.98 & 189.50 & 370.77 & 97.96 & 22.95 \\
\bottomrule
\end{tabular*}
\vspace{-0.05in}
\caption{This table shows the quantitative analysis of our method with other comparative methods on a synthetic dataset. Although our method is not the fastest in terms of speed, our method effectively trades off speed and accuracy.}
\label{tab:results1}
\vspace{-2mm}
\end{table*}

\begin{figure*}[t]\fontsize{1pt}{2pt}\selectfont
	\begin{center}
		\tabcolsep 1pt
		\begin{tabular}{@{}ccccccccc@{}}
			
			\includegraphics[width=0.095\textwidth]{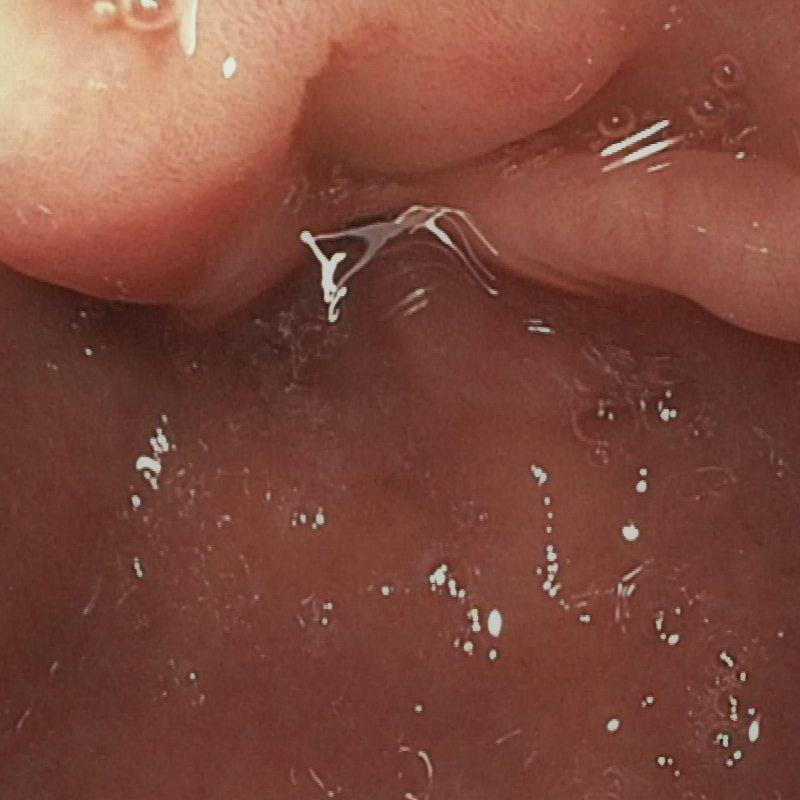} &
			\includegraphics[width=0.095\textwidth]{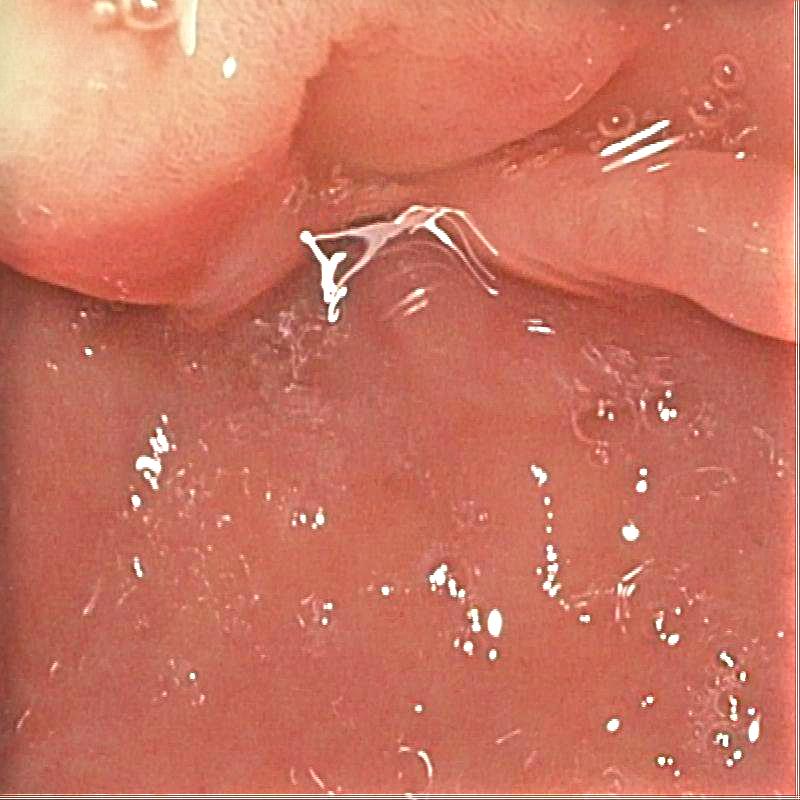} &
			\includegraphics[width=0.095\textwidth]{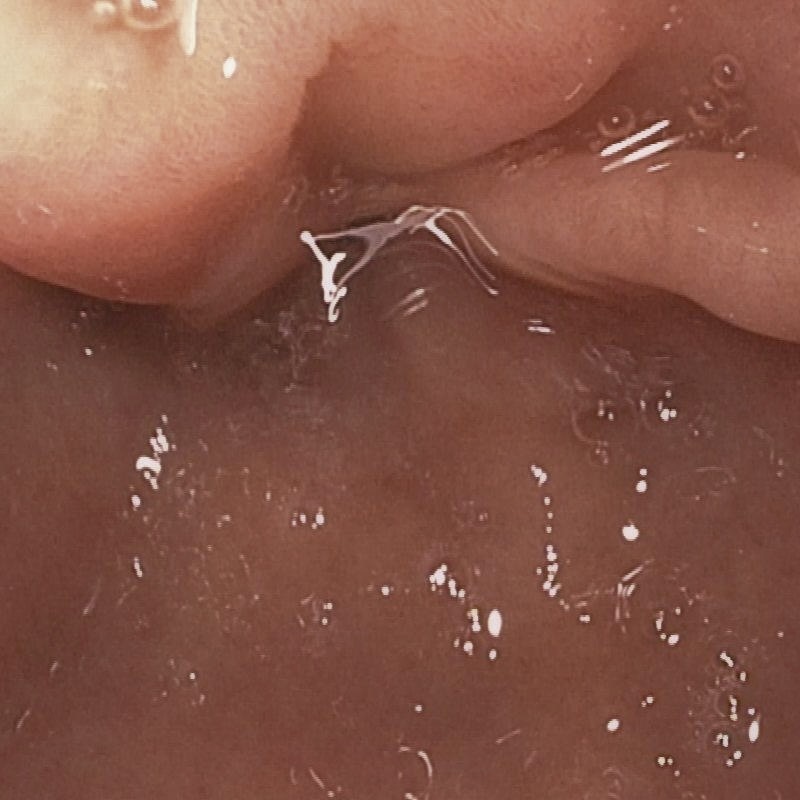} &
			\includegraphics[width=0.095\textwidth]{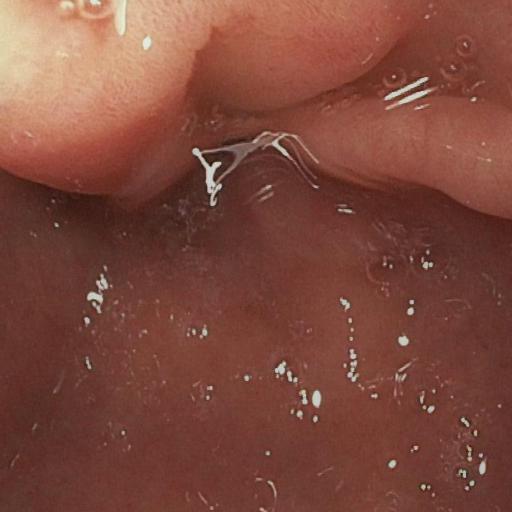} &
			\includegraphics[width=0.095\textwidth]{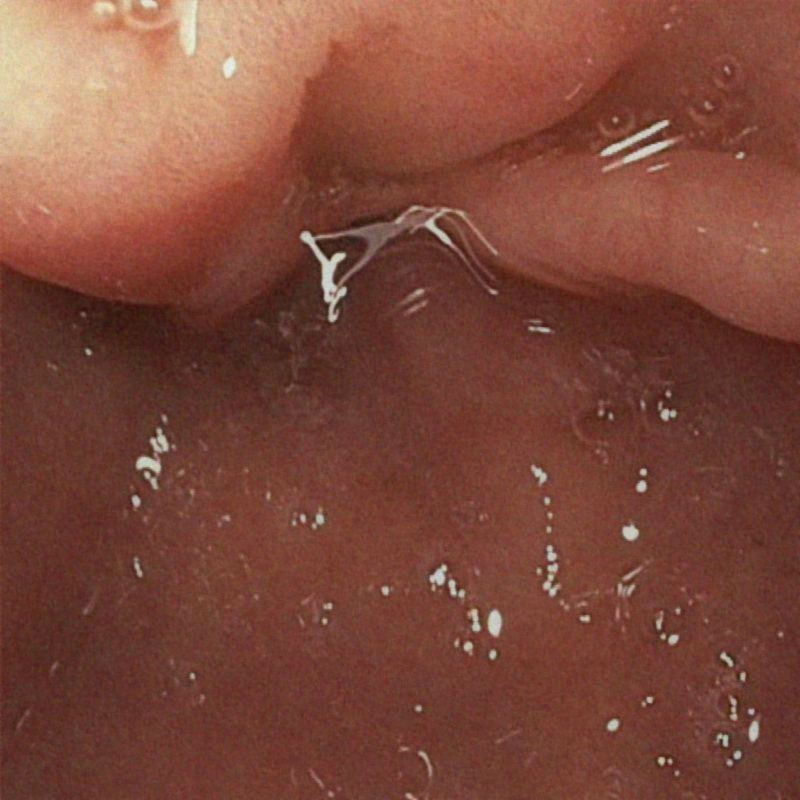} &
			\includegraphics[width=0.095\textwidth]{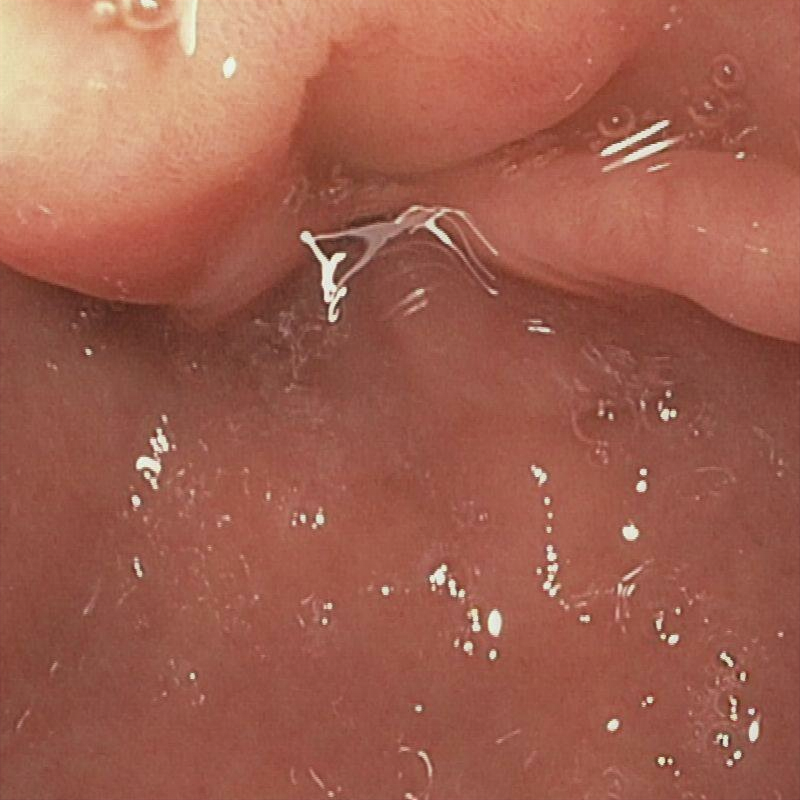} &
			\includegraphics[width=0.095\textwidth]{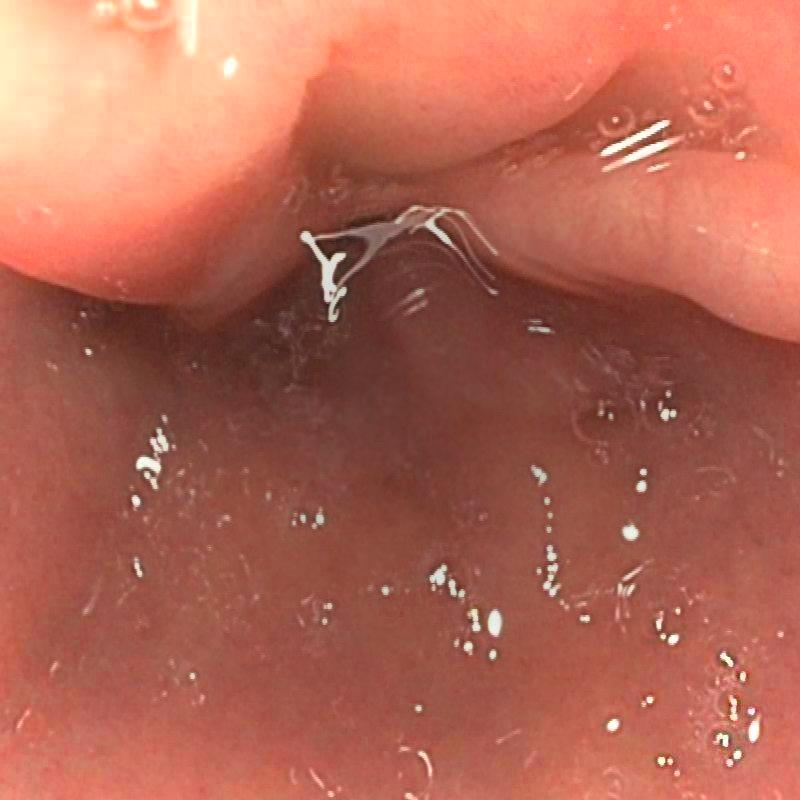} &
			\includegraphics[width=0.095\textwidth]{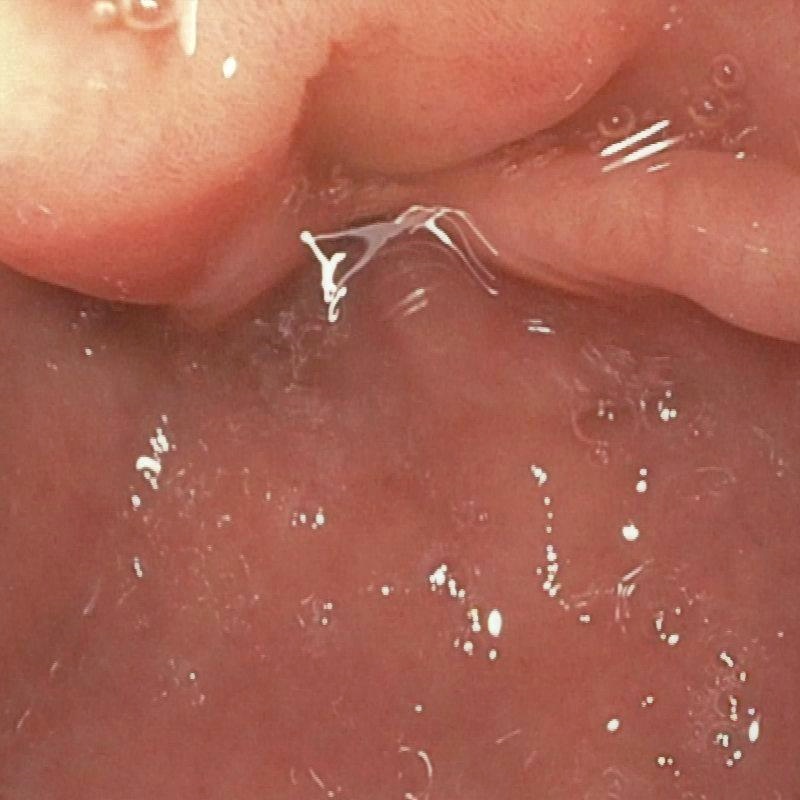} &
			\includegraphics[width=0.095\textwidth]{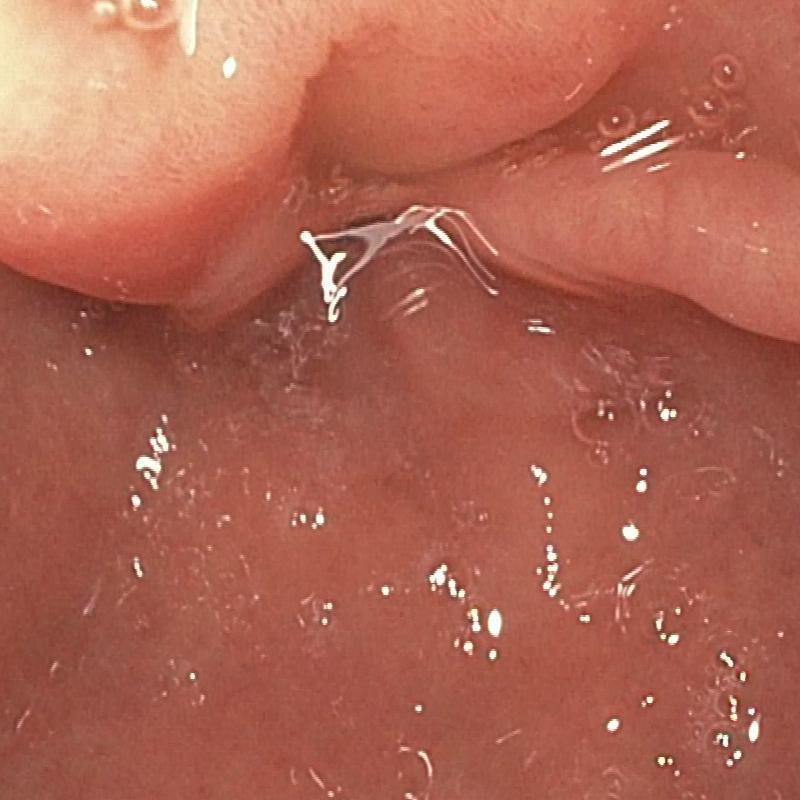} \\
			
			(PSNR/SSIM)  &
			(19.83/0.9198)  &  
			(16.25/0.7606)    & 
			(12.21/0.5074)   &
			(16.61/0.7913)     &
			\textbf{(29.40/0.9784)}     & 
			(20.68/0.9277)      &
			(27.19/0.9722)   &
			($+\infty$/1)     \\
			
			\includegraphics[width=0.095\textwidth]{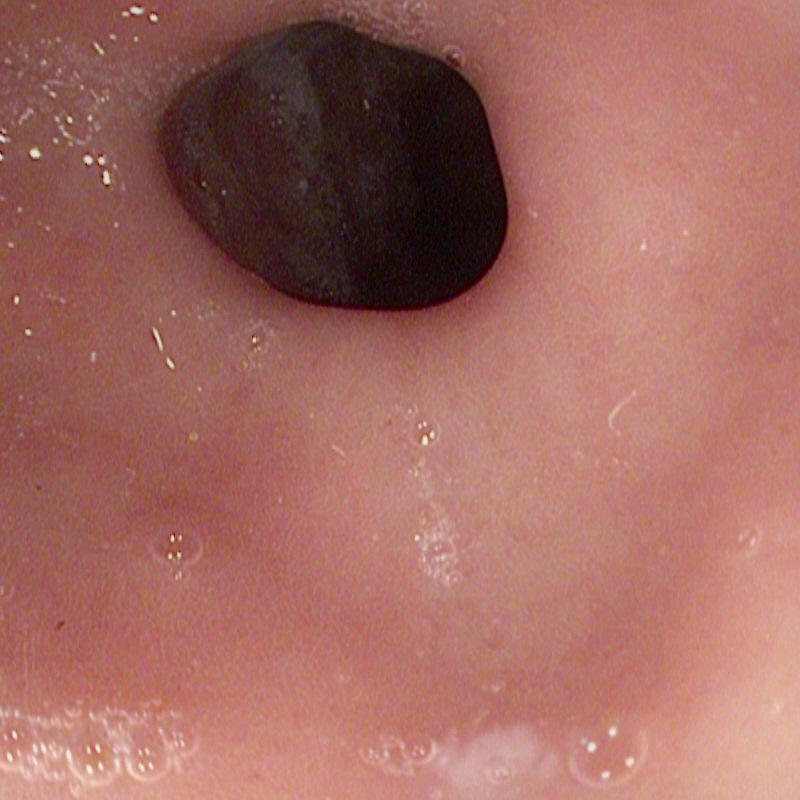} &
			\includegraphics[width=0.095\textwidth]{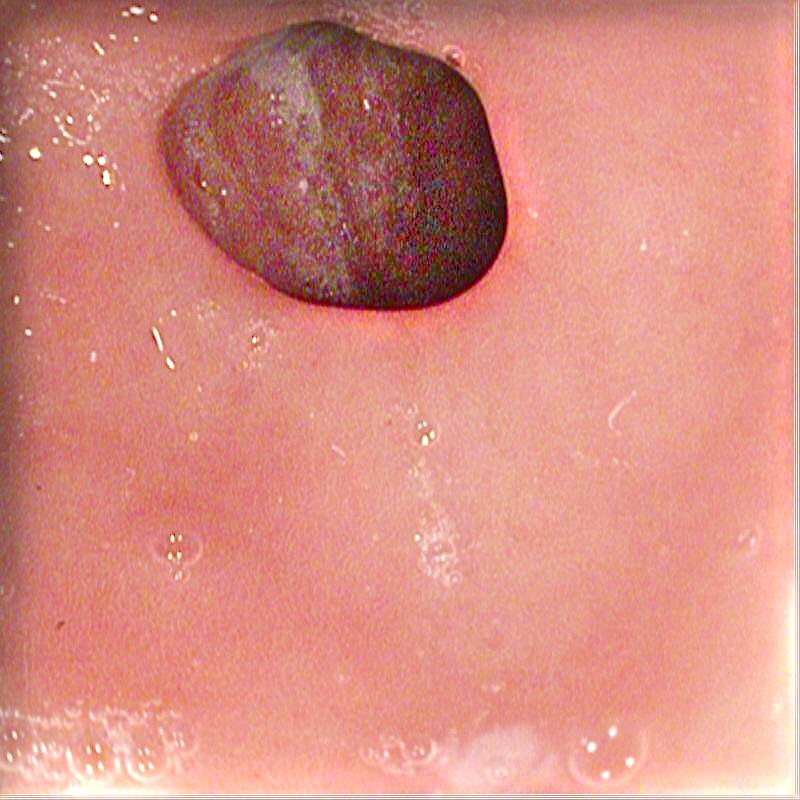} &
			\includegraphics[width=0.095\textwidth]{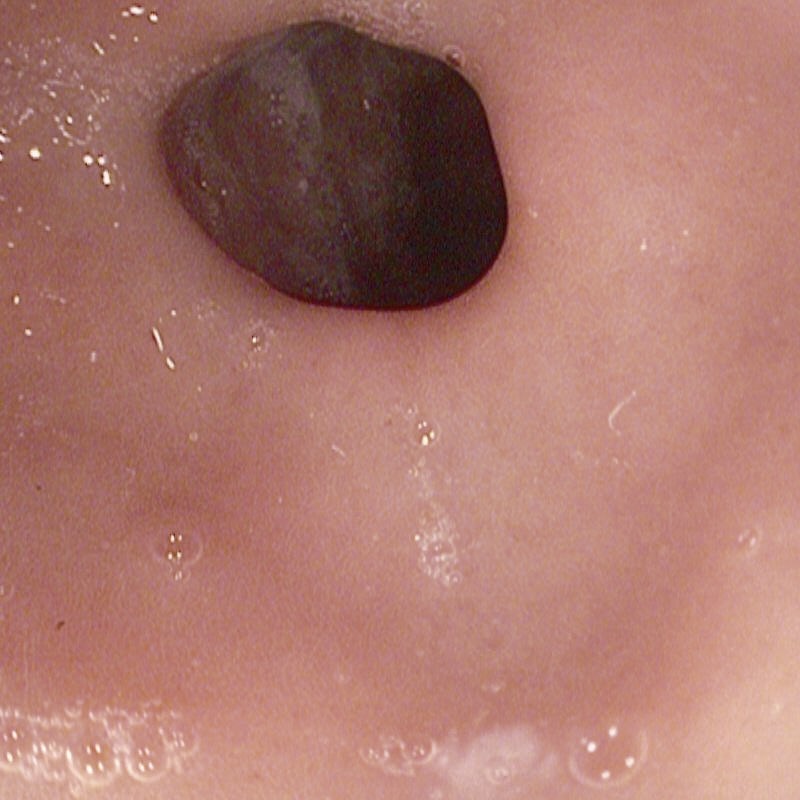} &
			\includegraphics[width=0.095\textwidth]{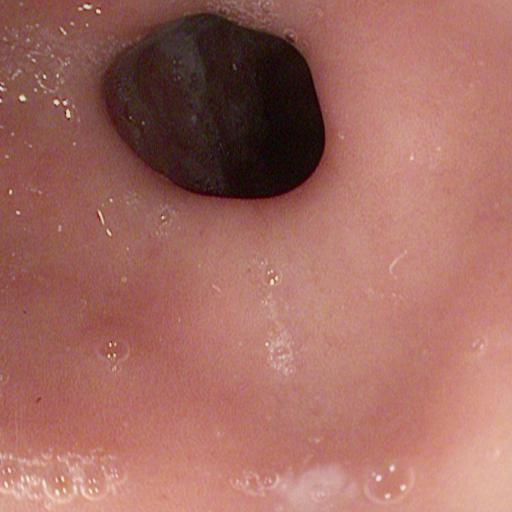} &
			\includegraphics[width=0.095\textwidth]{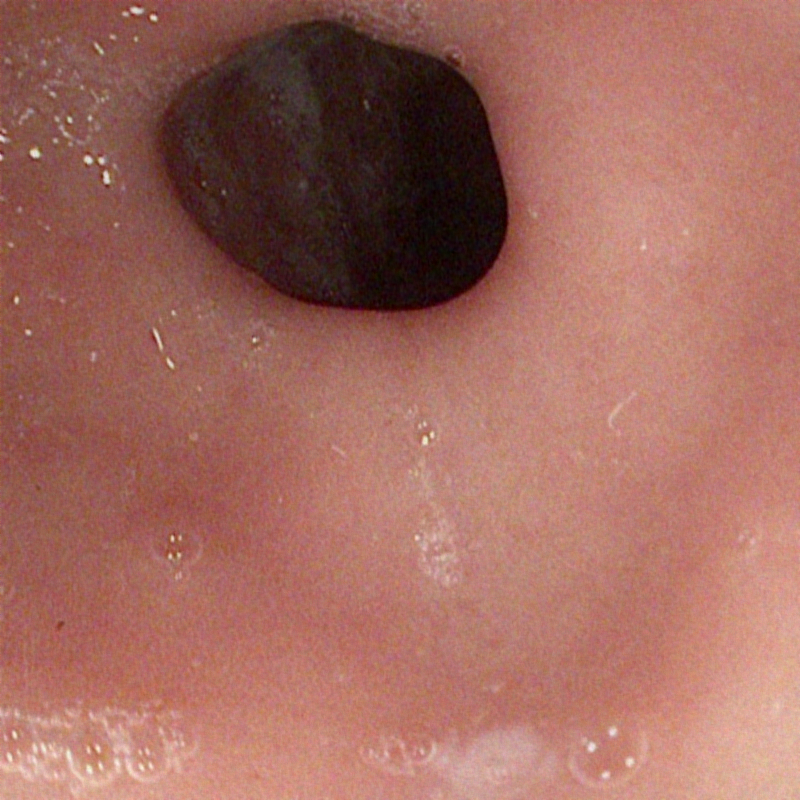} &
			\includegraphics[width=0.095\textwidth]{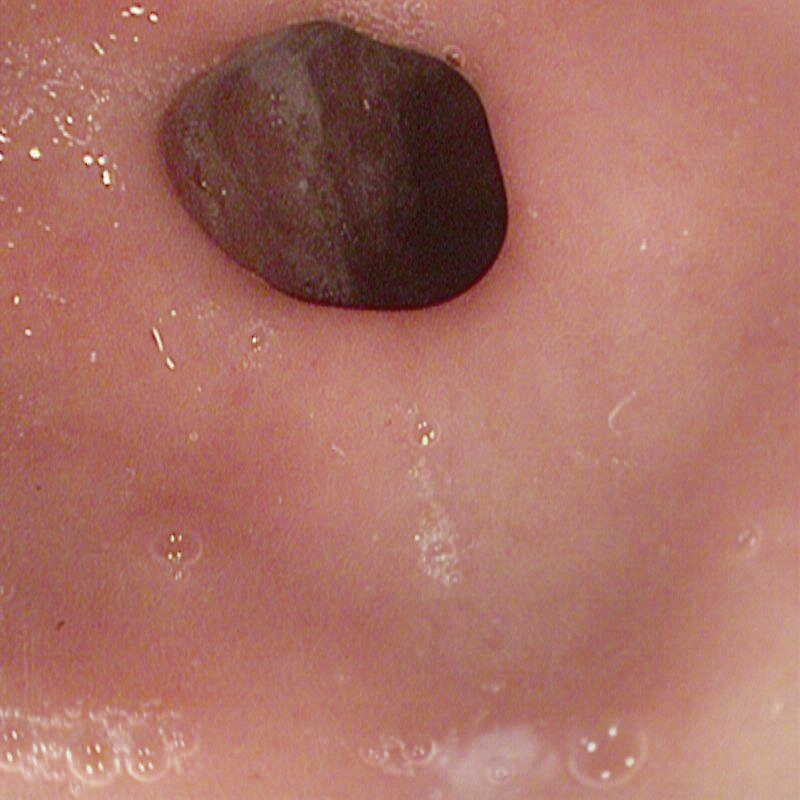} &
			\includegraphics[width=0.095\textwidth]{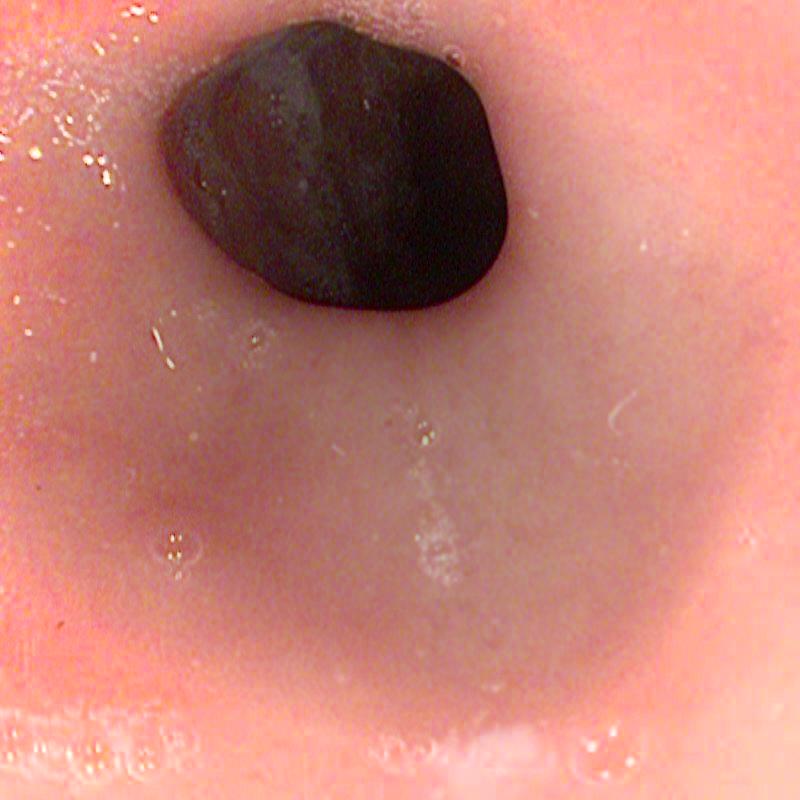} &
			\includegraphics[width=0.095\textwidth]{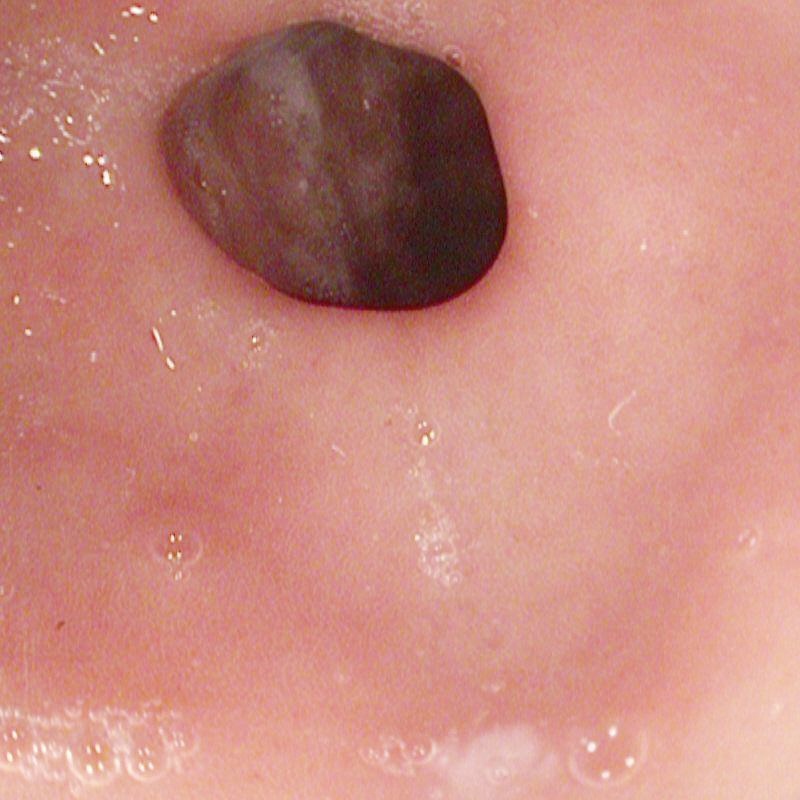} &
			\includegraphics[width=0.095\textwidth]{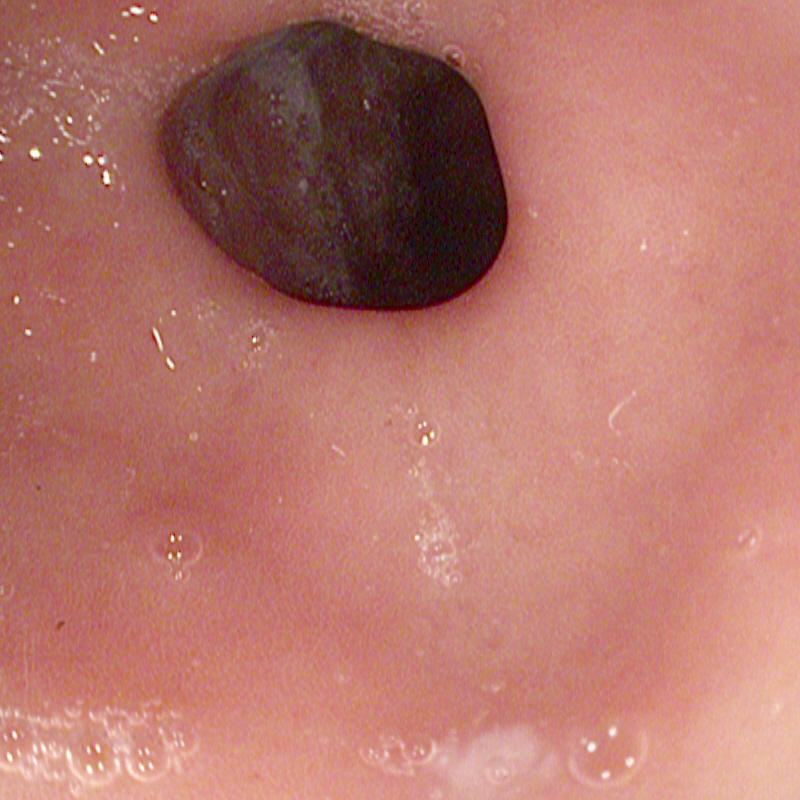} \\
			
			(PSNR/SSIM)  &
			(18.98/0.9041)  &  
			(24.08/0.7885)    & 
			(12.80/0.4619)   &
			(24.61/0.8472)     &
			(25.09/0.9838)     & 
			(20.23/0.9045)      &
			\textbf{(35.54/0.9821) }  &
			($+\infty$/1)     \\
			
			\includegraphics[width=0.095\textwidth]{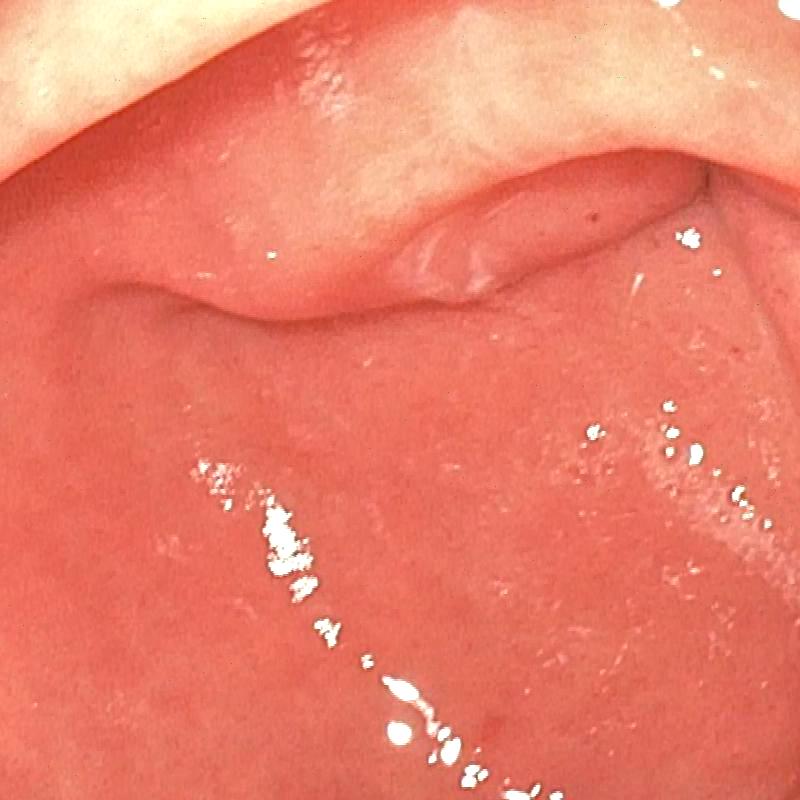} &
			\includegraphics[width=0.095\textwidth]{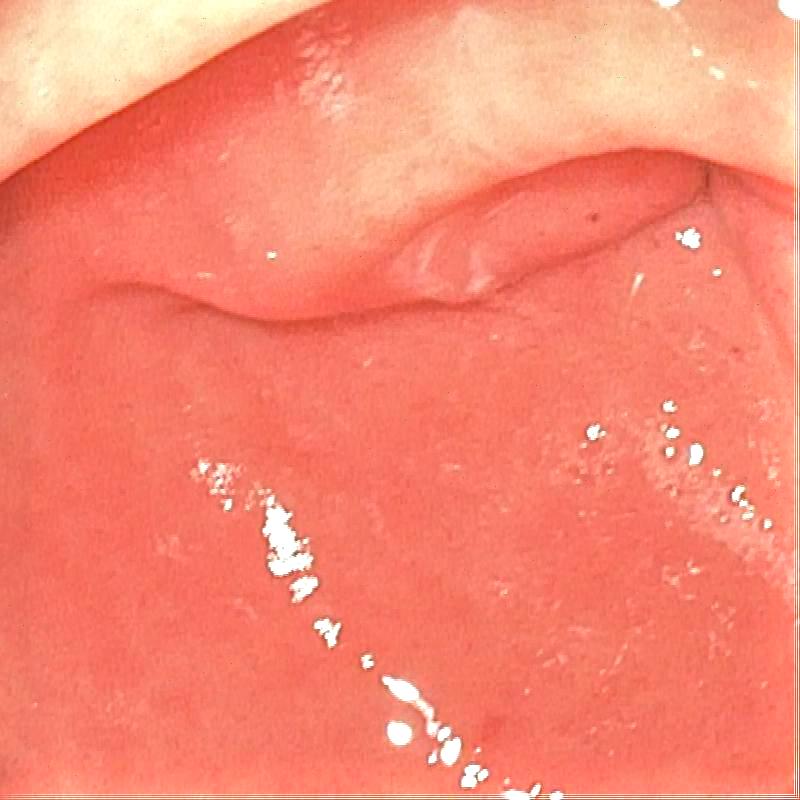} &
			\includegraphics[width=0.095\textwidth]{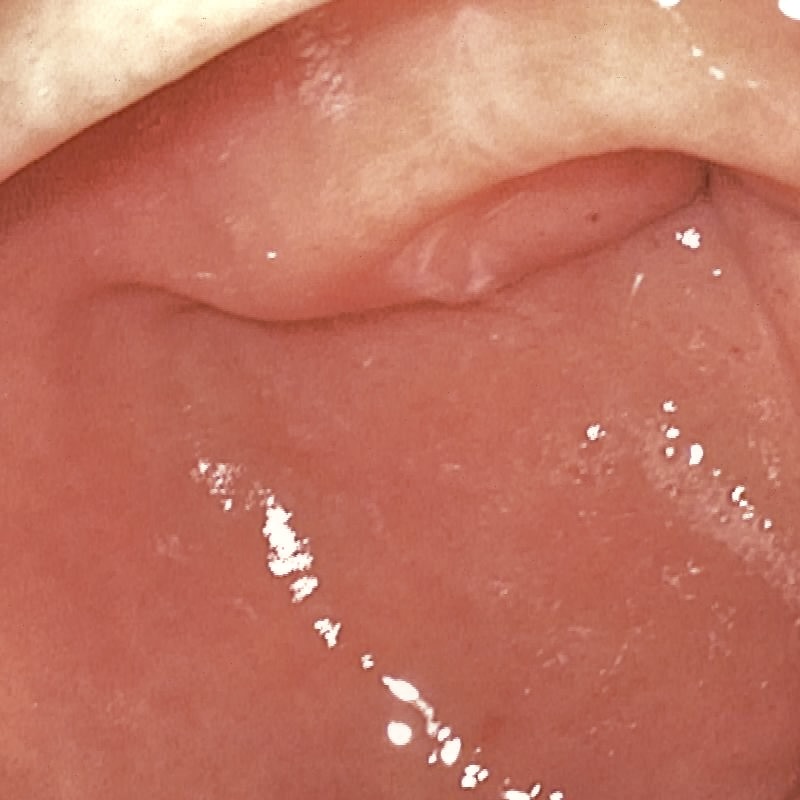} &
			\includegraphics[width=0.095\textwidth]{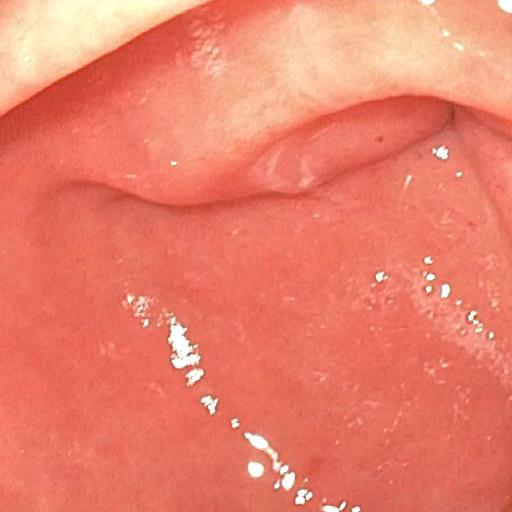} &
			\includegraphics[width=0.095\textwidth]{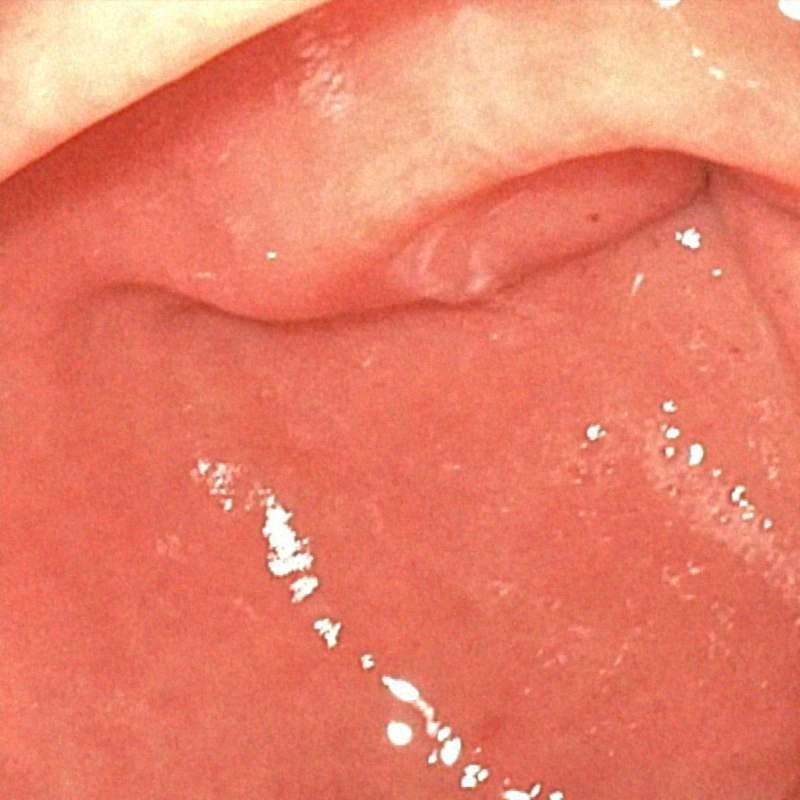} &
			\includegraphics[width=0.095\textwidth]{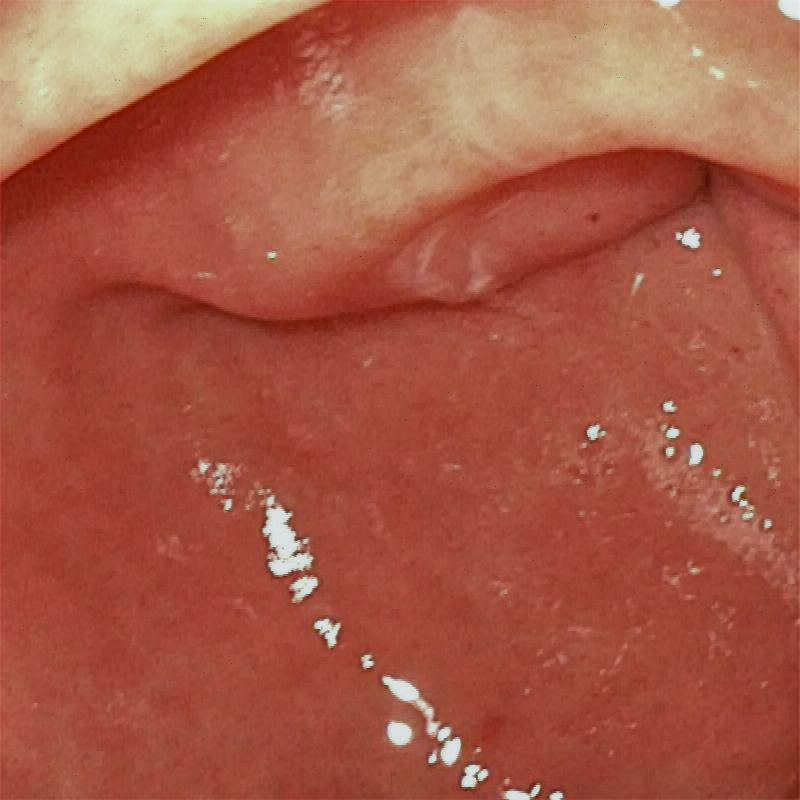} &
			\includegraphics[width=0.095\textwidth]{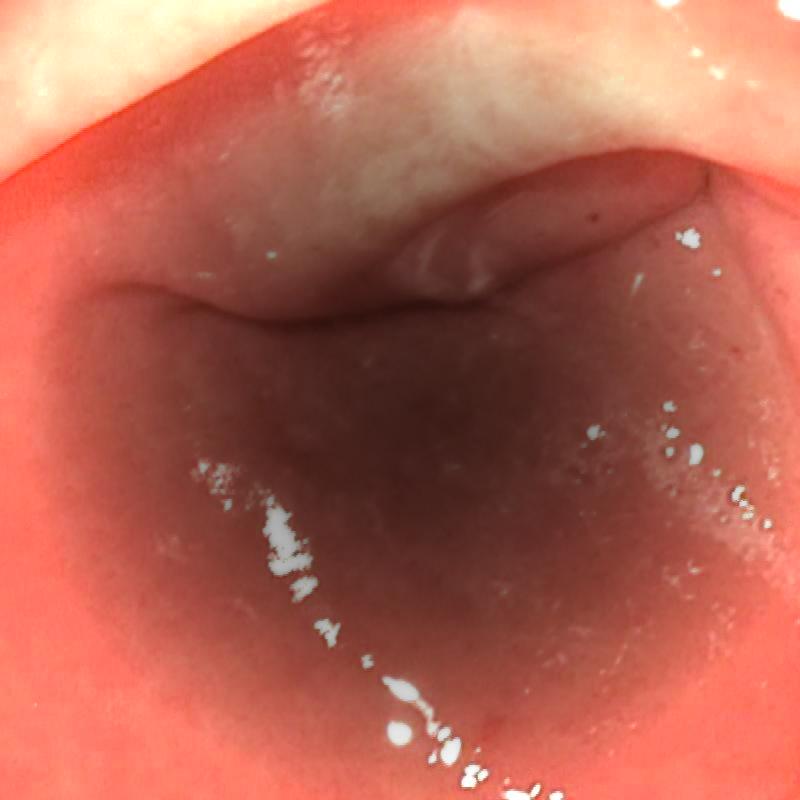} &
			\includegraphics[width=0.095\textwidth]{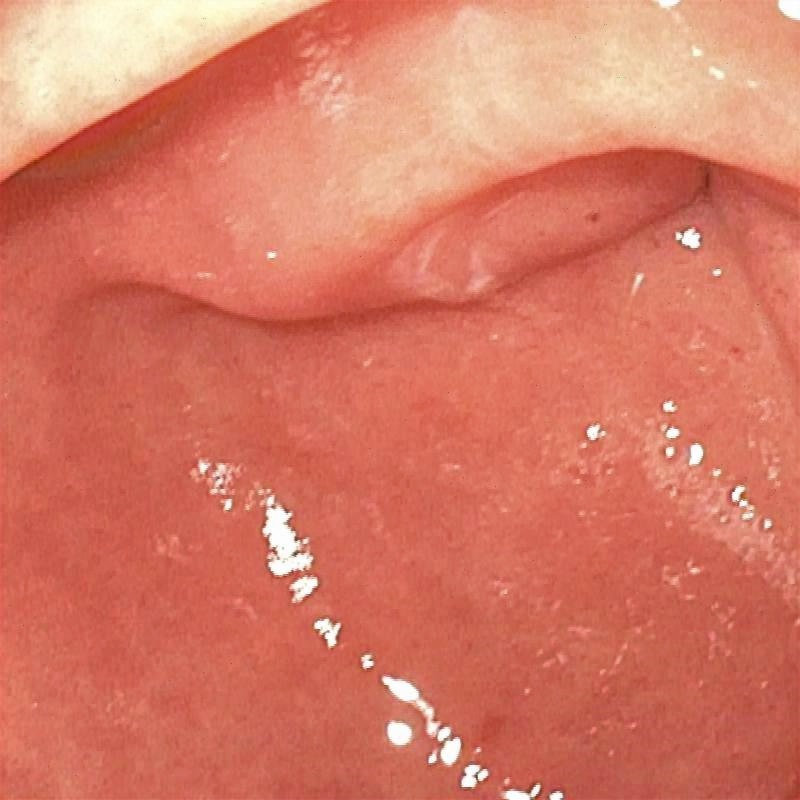} &
			\includegraphics[width=0.085\textwidth]{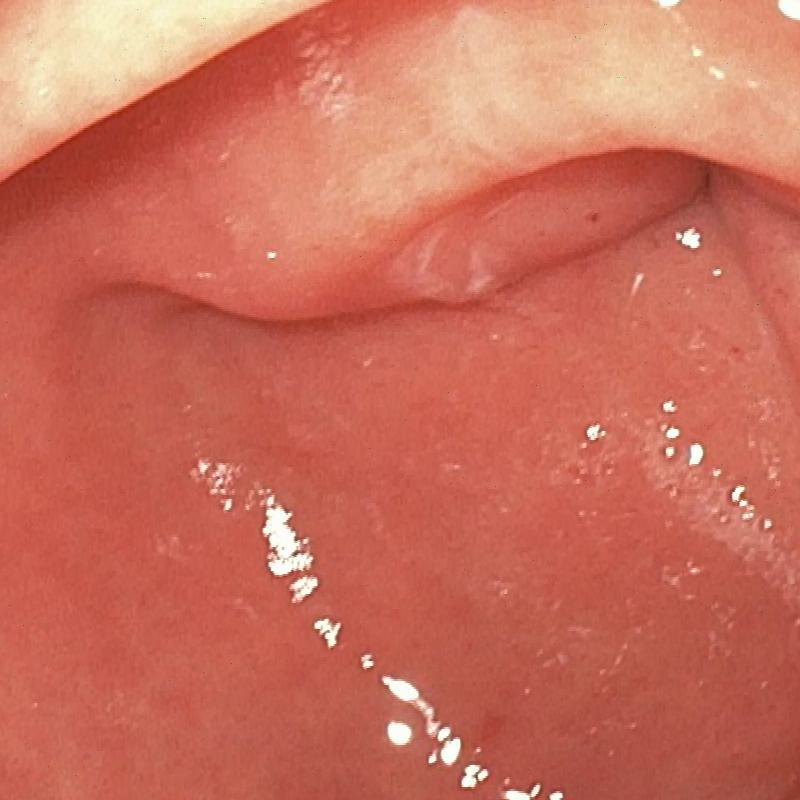} \\
			
			(PSNR/SSIM)  &
			(15.58/0.9433)  &  
			(18.70/0.8041)    & 
			(12.60/0.5067)   &
			(19.63/0.8192)     &
			(18.40/0.9594)     & 
			(14.96/0.8328)      &
			\textbf{(33.33/0.9812)}   &
			($+\infty$/1)     \\
			
			\includegraphics[width=0.095\textwidth]{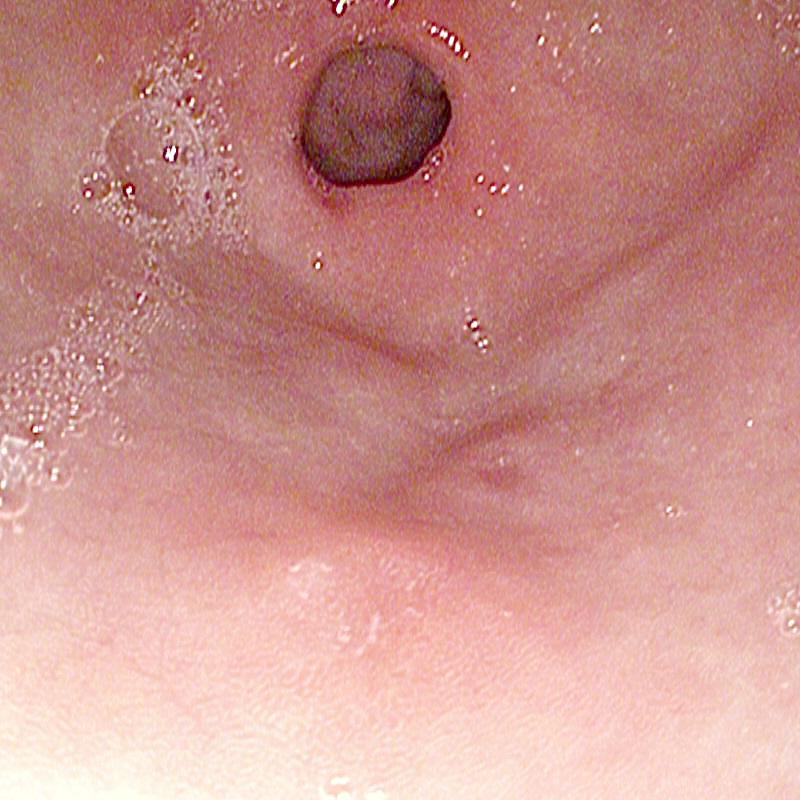} &
			\includegraphics[width=0.095\textwidth]{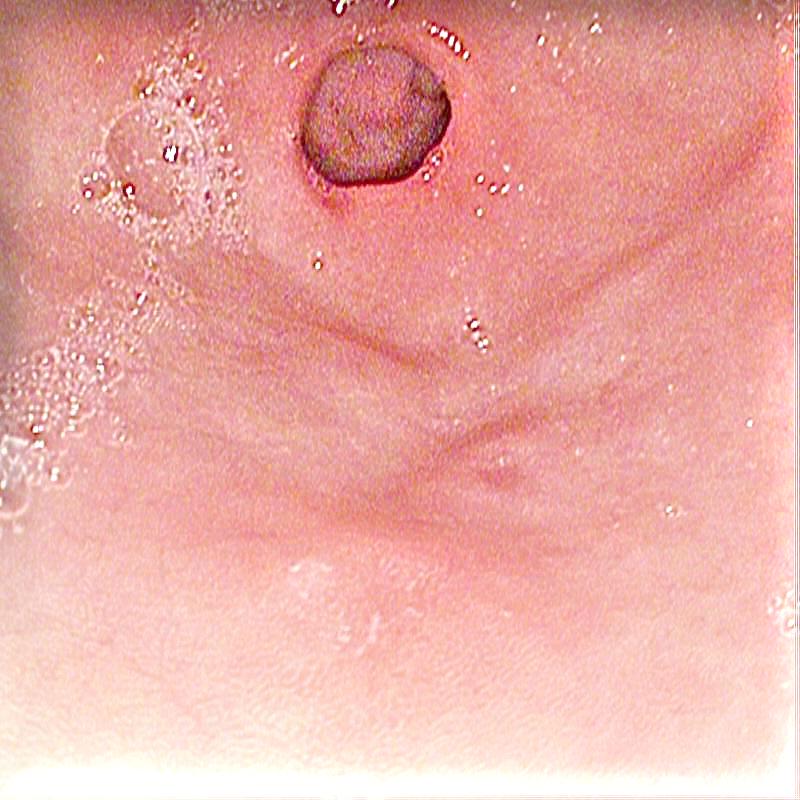} &
			\includegraphics[width=0.095\textwidth]{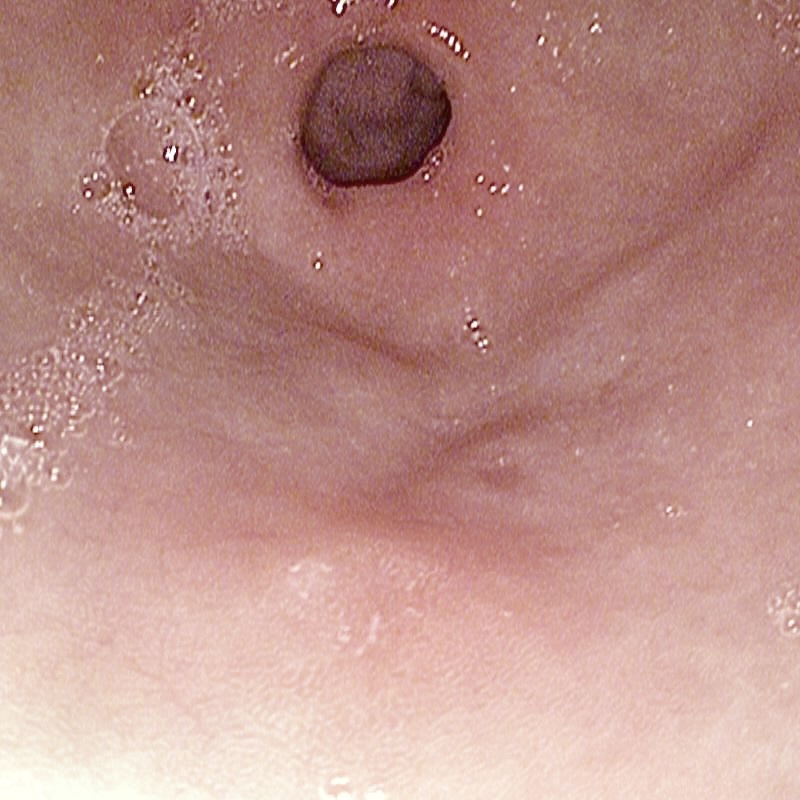} &
			\includegraphics[width=0.095\textwidth]{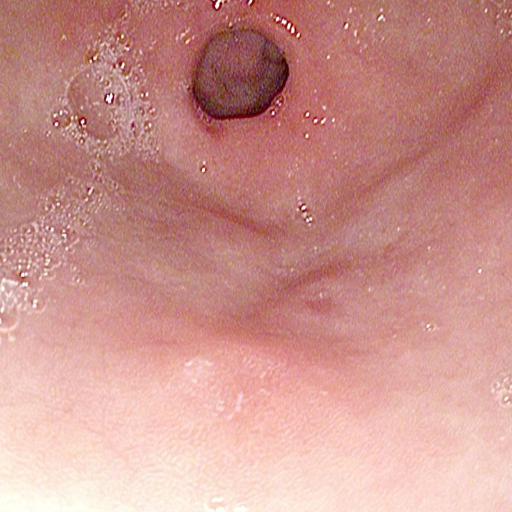} &
			\includegraphics[width=0.095\textwidth]{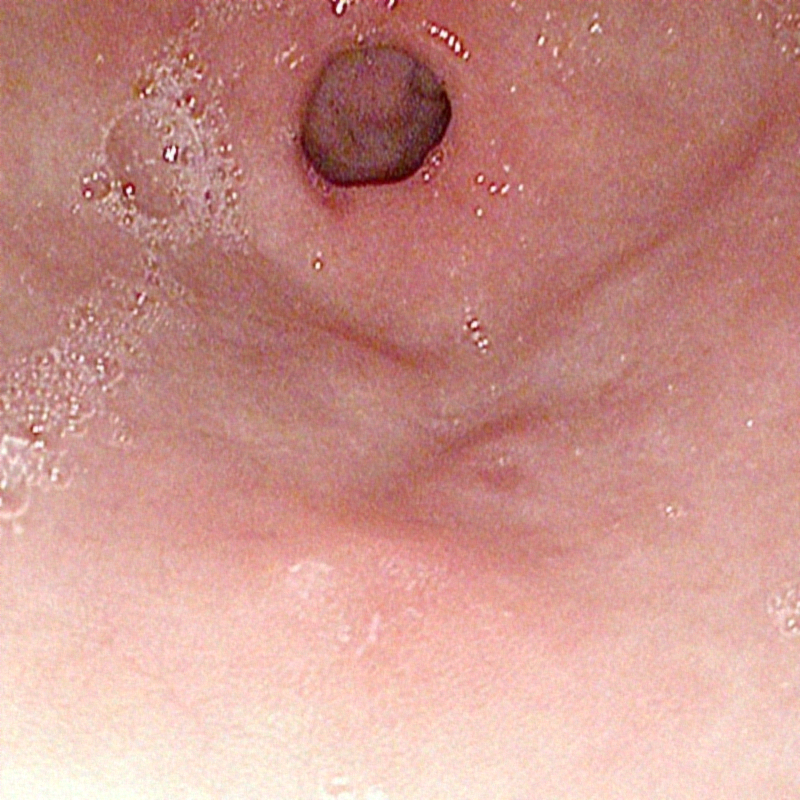} &
			\includegraphics[width=0.095\textwidth]{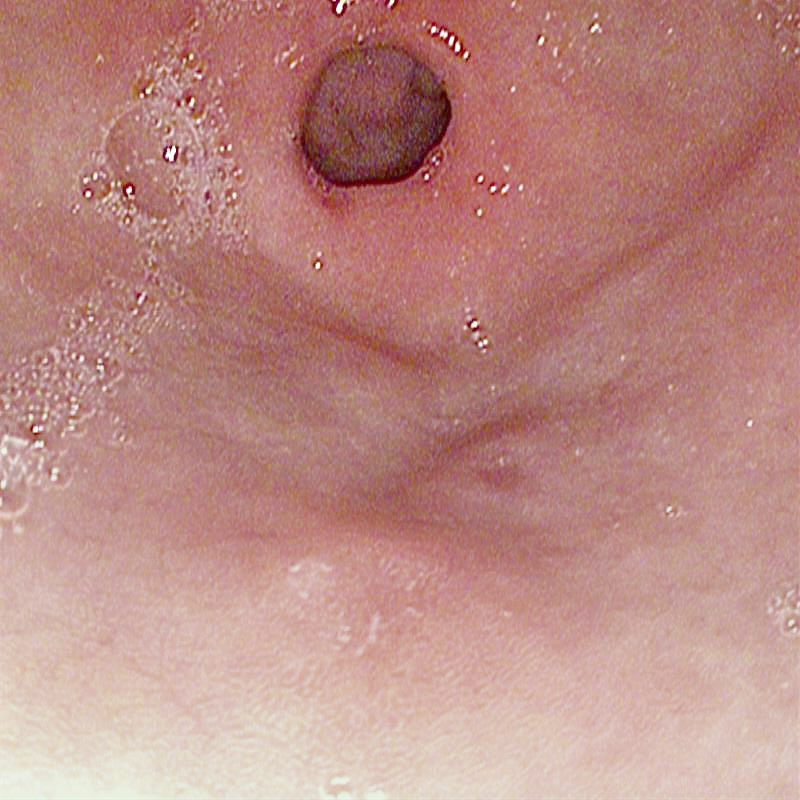} &
			\includegraphics[width=0.095\textwidth]{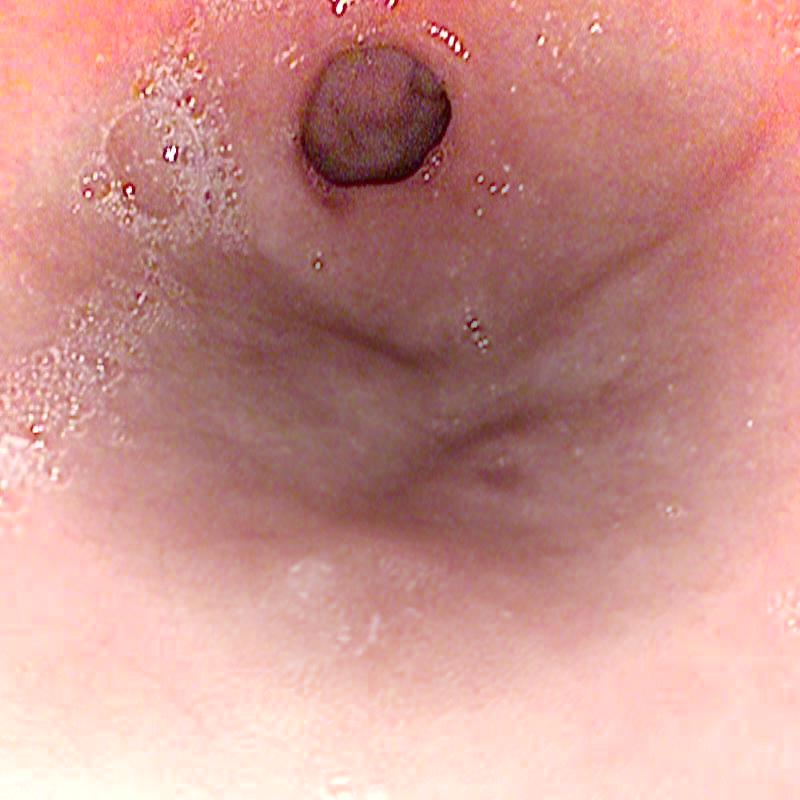} &
			\includegraphics[width=0.095\textwidth]{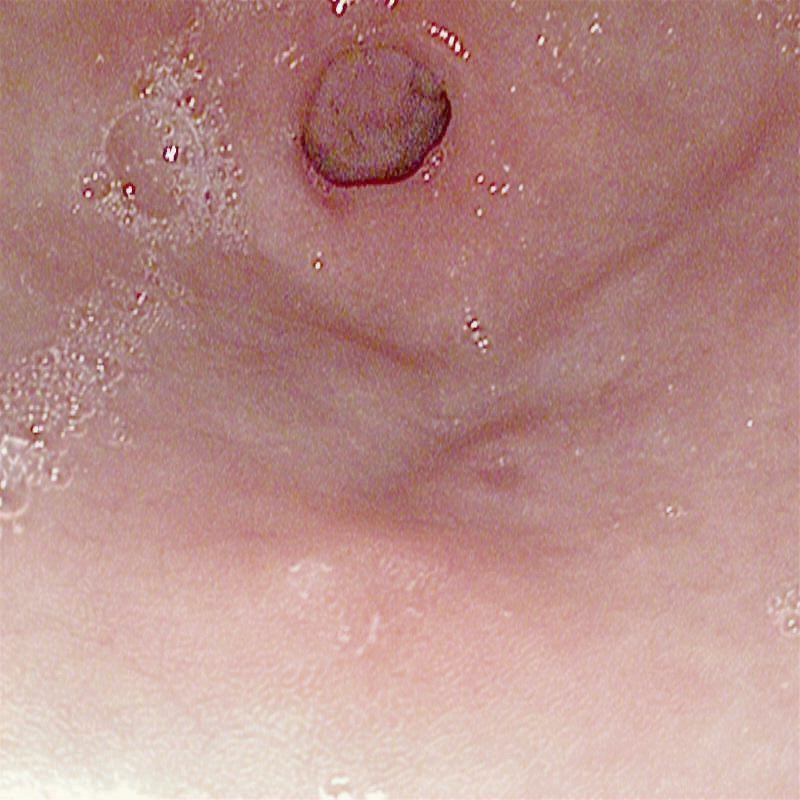} &
			\includegraphics[width=0.095\textwidth]{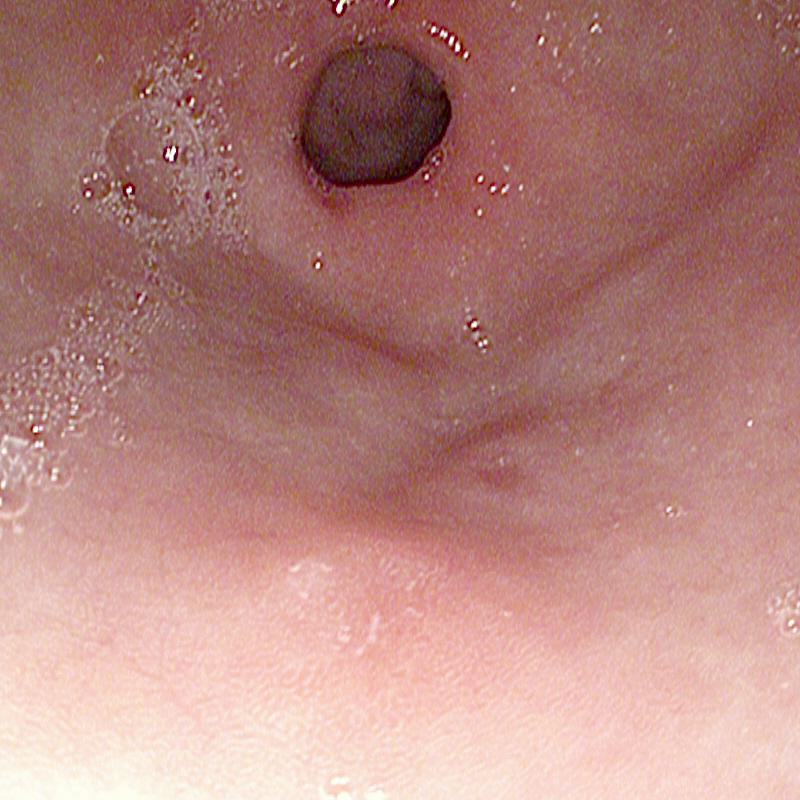} \\
			
			(PSNR/SSIM)     &
			(14.61/0.8990)  &  
			(21.51/0.6768)  & 
			(12.06/0.3324)  & 
			(22.40/0.8726)  & 
			(22.60/0.9787)        &  
			(18.20/0.8649)      & 
			\textbf{(27.64/0.9802)}   &
			($+\infty$/1)    \\
			
			(a) Input &
			(b) LIME &  
			(c) HDRNet & 
			(d) LECCM &
			(e) SwinIR &
			(f) NAFNet & 
			(g) EndolMLE &
			(h) Ours &
			(i) GT \\
			
		\end{tabular}
	\end{center}
	\vspace{-4mm}
	\caption{Exposure correction comparison on the E-kvasir. Our method outperforms other state-of-the-art techniques (\textit{LIME} \cite{guo2016lime}, \textit{HDRNET} \cite{gharbi2017deep}, \textit{LECCM} \cite{nsampi2021learning}, \textit{SwinIR} \cite{liang2021swinir}, \textit{NAFNet} \cite{chen2022simple} and \textit{EndoIMLE} \cite{wang2022endoscopic}), demonstrating enhanced visual quality and detail restoration.}
	
	\vspace{-4mm}
	\label{fig-Real_mydata}
\end{figure*}

\begin{table}[t] \scriptsize
\centering
\begin{tabular*}{0.5\textwidth}{@{\extracolsep{\fill}}cccc}
\toprule
& w/o cross-attention & w/o SSM  & Ours \\
\midrule
PSNR (dB) & 22.41 & 30.12  &  \textbf{33.99}  \\
SSIM & 0.8553 & 0.9530  & \textbf{0.9805} \\
\bottomrule
\end{tabular*}
\caption{Effectiveness of each module in our method.}
\label{tab:results2}
\end{table}

\section{Experiments}
\label{sec:experiments}
In this section, we evaluate the proposed approach through experiments on synthetic datasets and real-world images. 
The outcomes are compared with seven state-of-the-art HDR methods: \textit{LIME} \cite{guo2016lime}, \textit{HDRNET} \cite{gharbi2017deep}, \textit{LECCM} \cite{nsampi2021learning}, \textit{SwinIR} \cite{liang2021swinir}, \textit{NAFNet} \cite{chen2022simple} , \textit{EndoIMLE } \cite{wang2022endoscopic} , and EndoLMSPEC \cite{garcia2023multi}. Ablation studies are also conducted to substantiate the effectiveness of each module within our network. 

\subsection{Training Data}
We synthesized a novel endoscopic image HDR dataset (E-kvasri) with different exposures. This dataset comprises 900 pairs of under-/over-exposed images with their corresponding normally-exposed counterparts sourced from the kvasri~\cite{Pogorelov:2017:KMI:3083187.3083212} collection. The abnormal exposure variations were synthesized using the LECARM model \cite{ren2018lecarm} with a randomized exposure range between (-1, 1). Ultimately,  750 image pairs were arbitrarily allocated for training purposes. The remaining 150 image pairs were set aside for testing.
\begin{figure*}[t]\tiny
	\begin{center}
		\tabcolsep 1pt
		\begin{tabular}{@{}ccccc@{}}

        \includegraphics[width=0.20\textwidth]{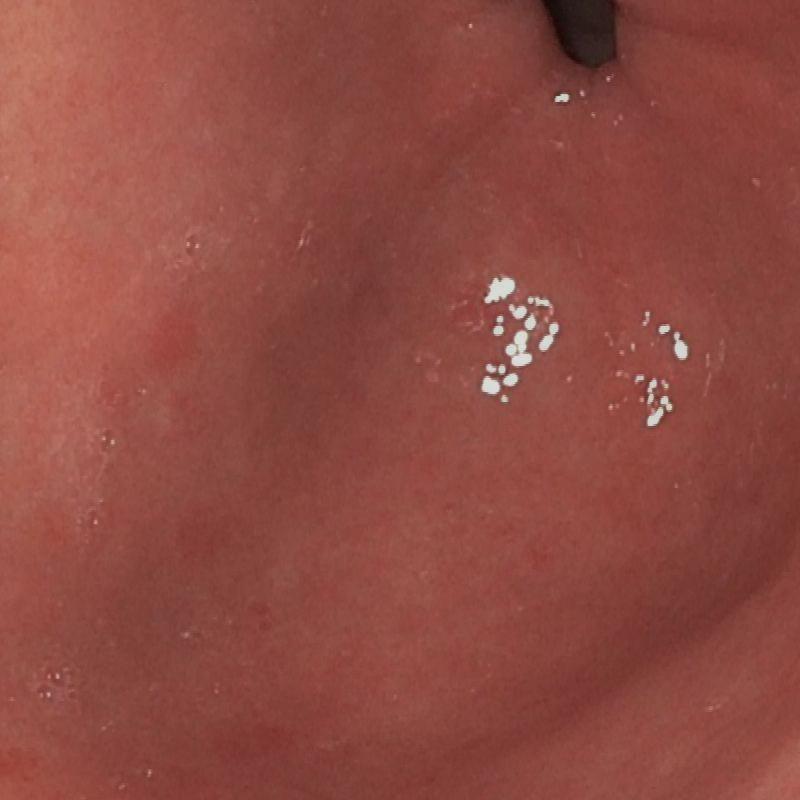} &
        \includegraphics[width=0.20\textwidth]{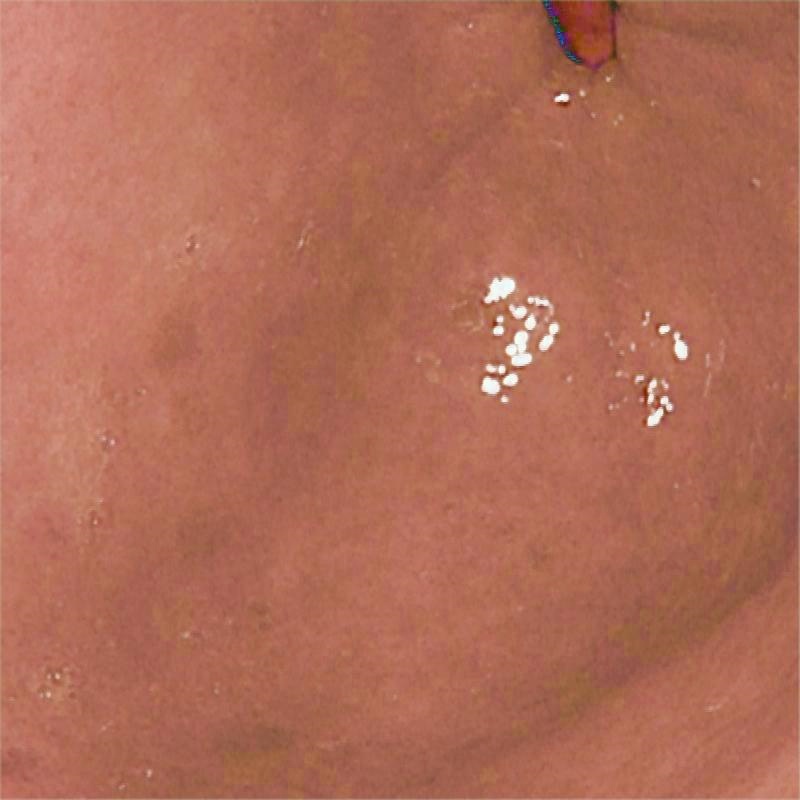} &
        \includegraphics[width=0.20\textwidth]{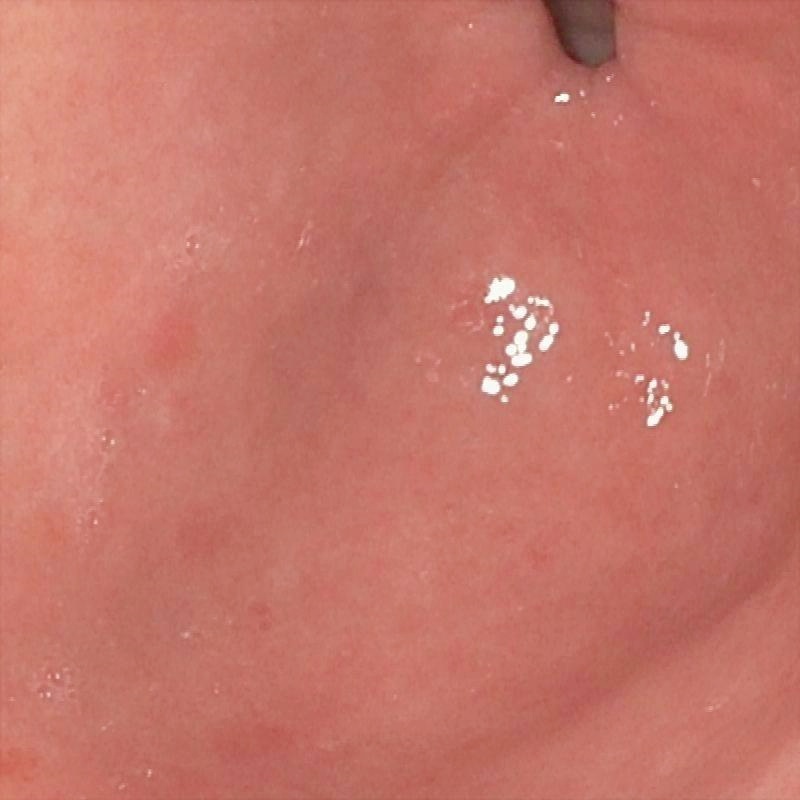} &
        \includegraphics[width=0.20\textwidth]{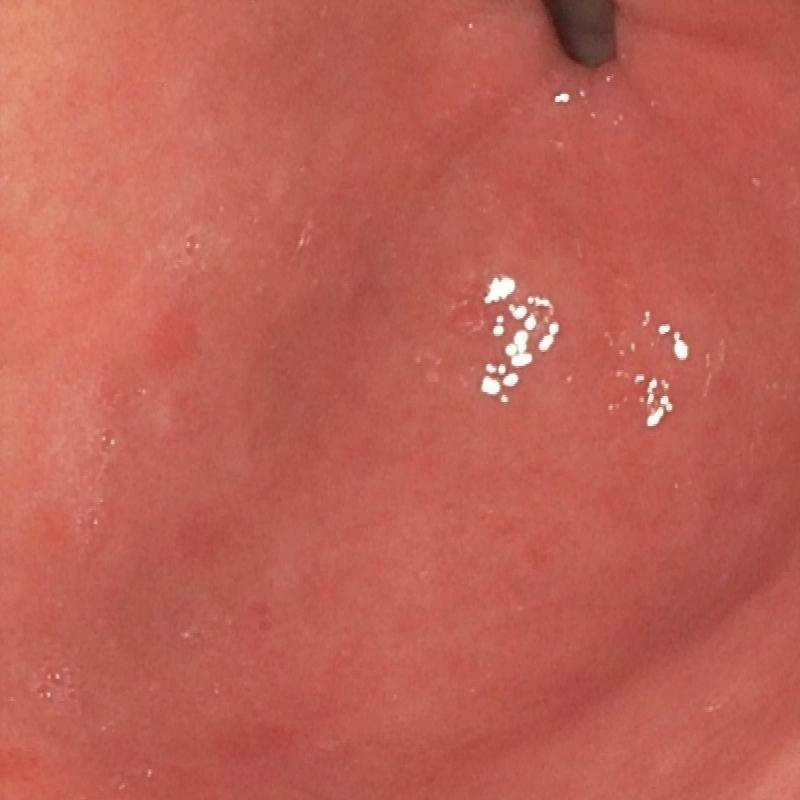} &
        \includegraphics[width=0.20\textwidth]{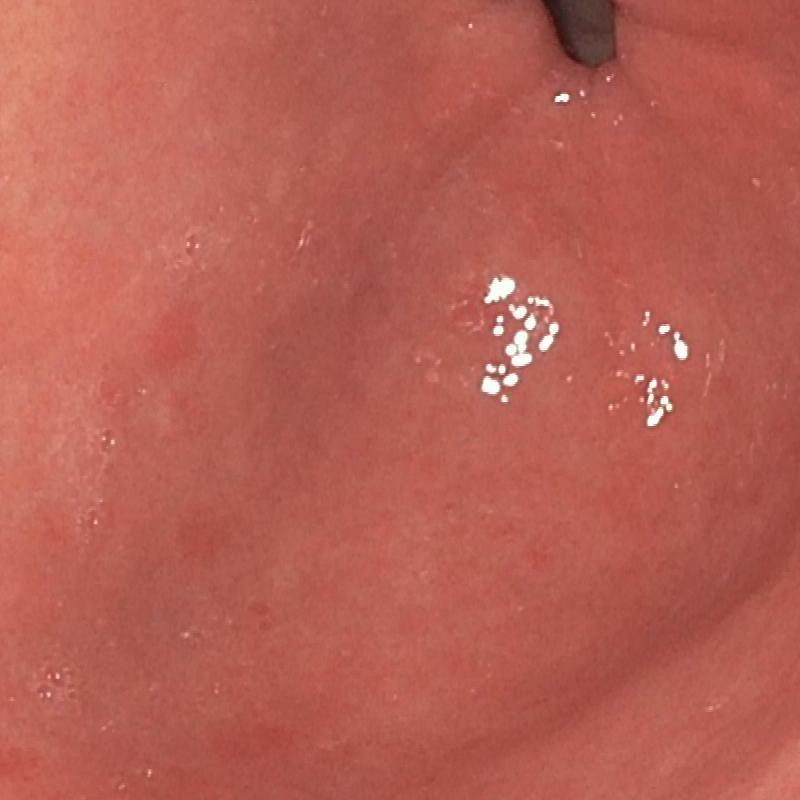} \\
        
        (32.11/0.9498)  & 
        (33.31/0.9510)  & 
        (35.33/0.9850)  & 
        \textbf{41.50/0.9899}  &  
        ($+\infty$/1)         \\

        (a) Input &
        (b) w/o Cross-attention& 
        (c) w/o SSM & 
        (e) Ours &  
        (f) GT         \\
		\end{tabular}
	\end{center}
	\vspace{-4mm}
	\caption{Comparison of exposure correction across individual modules of our method}
	\vspace{-4mm}
	\label{fig-ablations}
\end{figure*}

\subsection{Implementation details}
The proposed model was implemented in PyTorch 2.0 and trained using the Adam optimizer (a TITAN RTX3090 GPU shader with 24G RAM). We trained the network on image batches of size 32 with a resolution of \(512 \times 512\). The initial learning rate was set to 0.0002 with 600 epochs.

\subsection{Evaluation and Results}
The proposed method was evaluated on a synthetic dataset. All deep learning-based methods were fine-tuned based on E-kvasri. Figure~\ref{fig-Real_mydata} presents sample results of the proposed method and comparative methods on four endoscopic images from the E-kvasri.

\subsection{Ablations}

To validate the effectiveness of each module introduced in the proposed network, we conducted an ablation study involving the following two experiments:
1) Without Cross-attention: We removed cross-attention from the model.
2) Without SSM: We removed SSM from the model using a conventional linear layer in its place.
Table~\ref{tab:results2} and Figure~\ref{fig-ablations} compares our method with two ablation studies.


\section{Conclusion}
\label{sec:conclusion}
In this paper, we present a new architecture, FDVM-Net, for endoscopic image exposure correction, which integrates the advantages of local mode recognition of CNN and global context understanding of SSM. The results show that FDVM-Net has the promise to be the backbone of promising medical image enhancement networks for the next generation.

\bibliographystyle{splncs04}
\bibliography{ref}

\end{document}